\documentclass[letterpaper,journal]{IEEEtran}

\usepackage{cite}
\usepackage{amsmath,amssymb}
\usepackage{booktabs}
\usepackage{graphicx}
\usepackage{multirow}
\usepackage{threeparttable}
\usepackage{xcolor}
\usepackage{url}
\usepackage{xurl}
\usepackage[hidelinks]{hyperref}
\usepackage{tikz}
\usepackage{enumitem}
\usetikzlibrary{arrows.meta,positioning,fit,backgrounds,calc}

\newcommand{\model}{GigaSpeech~3}

\def\IEEEbibitemsep{-0.035\baselineskip}
\AddToHook{env/thebibliography/begin}{\label{ReferencesStart}}

\begin{document}
\bstctlcite{BSTcontrol}

\title{GigaSpeech 3: Scaling Multi-Dialect Arabic Speech Recognition with Weakly Labeled Data}

\author{%
Junzhe Liu\textsuperscript{1},
Yujie Tu\textsuperscript{2,3},
Yifan Yang\textsuperscript{1},
Peng Wang\textsuperscript{4},
Haoran Wang\textsuperscript{1},
Guodong Lin\textsuperscript{5},
Leyan Liao\textsuperscript{6},\\
Chengquan Zhang\textsuperscript{1},
Zhifu Gao\textsuperscript{7},
Xiangang Li\textsuperscript{7},
Qinyuan Cheng\textsuperscript{8},
Xipeng Qiu\textsuperscript{2,8}, \\
Wei-Qiang Zhang\textsuperscript{5,9},
Guoguo Chen\textsuperscript{9,10},
Kai Yu\textsuperscript{1},
Xie Chen\textsuperscript{1,2,9}\\[0.45em]
\small
\textsuperscript{1}Shanghai Jiao Tong University \quad
\textsuperscript{2}Shanghai Innovation Institute\\
\textsuperscript{3}University of Chinese Academy of Sciences \quad
\textsuperscript{4}The Chinese University of Hong Kong, Shenzhen\\
\textsuperscript{5}Tsinghua University \quad
\textsuperscript{6}Dalian University of Technology \quad
\textsuperscript{7}Alibaba Token Foundry\\
\textsuperscript{8}Fudan University \quad
\textsuperscript{9}SpeechColab \quad
\textsuperscript{10}Seasalt.ai
\thanks{\parbox[t]{0.95\columnwidth}{%
\noindent
This work has been submitted to the IEEE for possible publication. Copyright may be transferred without notice, after which this version may no longer be accessible.\par\smallskip
Code: \url{https://github.com/Antares-97/GigaSpeech3}\\
Dataset: \url{https://huggingface.co/datasets/speechcolab/gigaspeech3}\\
Demo: \url{https://chenxie95-gigaspeech3-arabic-asr.hf.space}\\
Proofreading: \url{https://open-ai-eval.com/}%
}}
}

\maketitle

\begin{abstract}
Dialectal Arabic speech recognition remains constrained by the scarcity of dialect-resolved transcribed speech despite the abundance of web audio. We introduce \model, a corpus of 39,000 hours and 10 million weakly transcribed speech segments covering Algerian, Egyptian, Iraqi, Saudi, and Syrian Arabic. Starting from 60 hours of manual transcription per dialect, we train labeling models using speech-recognition-biased self-supervised learning and apply them to dialect-matched web audio. Data selection and four rounds of noisy-student refinement for Saudi and Syrian Arabic yield the final corpus. We validate its utility through dialect-specific and multi-dialect speech recognition training and cross-domain evaluation. On GigaSpeechBench, the final dialect-specific models outperform their labeling models across all five dialects and achieve a macro-average word error rate of 23.86\%, a 16.6\% relative reduction from the strongest third-party system covering all five dialects, while outperforming the best third-party result on three dialects. A multi-dialect model using self-predicted dialect identities achieves a macro-average word error rate of 25.51\% while also supporting Modern Standard Arabic. Across eight conditions from three external corpora, a \model-trained Zipformer achieves the lowest word error rate among six evaluated models in every condition, while adapting the Qwen3 speech recognizer on \model{} improves performance across all eight conditions. We release the corpus and recipes as a foundation for dialect-aware Arabic speech processing.
\end{abstract}


\section{Introduction}
\label{sec:introduction}

Arabic ASR is a multi-variety problem rather than a single-language recognition setting. Modern Standard Arabic (MSA) is standardized and dominates education, news, and formal communication, whereas everyday speech consists of regional varieties with substantial differences in pronunciation, vocabulary, morphology, and spelling. A common coarse taxonomy distinguishes Maghrebi, Egyptian, Levantine, Gulf, and Iraqi or Mesopotamian Arabic, although boundaries vary and Iraqi is sometimes grouped with Gulf Arabic \cite{zaidan_arabic_2014,djanibekov_dialectal_2025}. Our five targets provide one country-level representative of each group: Algerian (DZA) for Maghrebi, Egyptian (EGY), Syrian (SYR) for Levantine, Saudi (SAU) for Gulf, and Iraqi (IRQ).

Arabic speech resources have developed along several directions. Challenge and benchmark datasets, including MGB-2, MGB-3, MGB-5, the Open Universal Arabic ASR Leaderboard, and GigaSpeechBench, have expanded evaluation across countries, domains, and recording conditions \cite{ali_mgb2_2016,ali_mgb3_2017,ali_mgb-5_2019,wang_open_2024,tu_gigaspeechbench_2026}. Large broadcast corpora such as QASR and SADA provide substantially more transcribed audio, but their coverage is shaped by Al Jazeera programming or Saudi television and therefore does not yield balanced country-level supervision for the five dialects studied here \cite{mubarak_qasr_2021,alharbi_sada_2024}. Carefully annotated corpora such as Casablanca, CARDAMOM, and Hamsa provide country-level or finer dialect labels, but their scale of tens of hours primarily supports evaluation, adaptation, and linguistic analysis rather than the data-intensive training targeted here \cite{talafha_casablanca_2024,talafha_cardamom_2026,alyafeai_hamsa_2026}. At the other extreme, web-scale collections make much more Arabic audio available: YODAS v3 reports roughly 25,000 hours of Arabic within a 1.1-million-hour multilingual corpus, yet its locale metadata are incomplete, and its Arabic partition does not provide reliable country-level dialect labels \cite{chen_yodas3_2026}. The remaining gap is therefore not the availability of Arabic speech data, but the lack of large-scale, dialect-specific training data with reliable dialect labels.

Preserving dialect identity matters because a pooled ``Arabic'' label hides systematic acoustic and lexical differences. We retain country-level labels through metadata-based collection and curation, then use them to organize weak supervision for dialect-specific and multi-dialect recognizers. The multi-dialect recognizer uses automatically predicted dialect IDs at inference. A matched comparison with known IDs shows that routing errors increase WER, especially for EGY and SYR. We therefore evaluate this practical cost while keeping improvements to dialect-identification methodology outside the claimed contribution.

Large-scale self-supervised pretraining offers one route around the transcription bottleneck. Methods such as wav2vec~2.0, HuBERT, and XLS-R learn speech representations from untranscribed audio and adapt them with much smaller labeled sets \cite{baevski_wav2vec_2020,hsu_hubert_2021,babu_xls-r_2021}. Multilingual pretraining can further share statistical strength across languages \cite{pratap_massively_2020,ramirez_anatomy_2024}, but a general multilingual model does not create dialect-specific transcripts for training a compact recognizer. In practice, dialect-matched web audio may be plentiful while the supervision needed to train Arabic dialect ASR remains scarce.

Automatic weak labeling turns this asymmetry into a corpus-construction strategy. A labeling model transcribes unlabeled speech, and a downstream model learns from the resulting weakly labeled audio--text pairs. Iterative labeling and related noisy-student methods have produced large gains in low-resource and multilingual ASR \cite{xu_iterative_2020,lugosch_pseudo-labeling_2022,bhogale_empowering_2024, kang2024libriheavy}. GigaSpeech~2 extends this idea into an end-to-end data pipeline that crawls web audio, generates transcripts, filters unreliable pairs, and iteratively refines the labels \cite{yang_gigaspeech_2024}. For Arabic dialects, however, the decisive question is whether automatically generated transcripts are accurate and diverse enough to become a reusable training corpus. A strong test is whether an independently initialized downstream model can improve over the labeling model without inheriting its pretrained parameters.

We present \model, a large-scale weakly labeled speech corpus for the five target dialects. We partition web speech by dialect and keep labeled, unlabeled, and test channels disjoint. Starting from a small manually transcribed set per dialect, the ASR-biased self-supervised recipe introduced by VietASR produces labeling models for the larger matched collections \cite{zhuo2025vietasr}. Selection and NST refinement yield 39,028 hours and 10 million segments, with 4,361--11,461 hours per dialect. The released SAU/SYR partitions are filtered through the fourth NST round; their corresponding final recognizers are denoted $r_4$ (NST round 4 models). We validate the weak supervision through new dialect-specific ASR models; a multi-dialect model pools the dialect inventories with separately labeled MSA data and uses predicted dialect IDs at inference.
We also provide an online proofreading platform for community contributors to review and correct weak transcriptions, supporting future corpus improvements.

The central result is that \model\ provides useful supervision beyond the labeling models that generated it. Final dialect-specific models, including NST r4 for SAU/SYR, improve over their corresponding labeling models on all five dialects and reach a 23.86 macro-average WER on GigaSpeechBench. The multi-dialect model reaches 25.51 with predicted LID, compared with 28.62 for the strongest complete third-party system. The dialect-specific models lead on DZA, EGY, and SAU; the strongest third-party system retains an advantage on IRQ and SYR. On eight external conditions from MASC, Casablanca, and SADA, a \model-trained Zipformer obtains the lowest WER among the six evaluated models in every condition. Corpus-based Qwen3-ASR fine-tuning also improves it on all eight, supporting transfer across evaluation domains and model families.




Our main contributions are summarized as follows:
\begin{itemize}[leftmargin=*, itemsep=2pt, topsep=2pt]
\item \textbf{Large-scale dialect-resolved corpus.}
We introduce \model, comprising 39,028 hours and 10 million weakly transcribed segments across five Arabic dialects (DZA, EGY, IRQ, SAU, and SYR), bridging the gap between small manually annotated dialect corpora and large Arabic speech collections lacking reliable dialect labels.

\item \textbf{Validation of weak supervision.}
Final dialect-specific models outperform their labeling models across all five dialects, achieving a macro-average WER of 23.86 and a 16.6\% relative reduction over the strongest third-party system covering all five dialects on GigaSpeechBench.

\item \textbf{Evaluation across settings and domains.}
We evaluate multi-dialect recognition with predicted-LID decoding and MSA coverage, quantify the WER degradation caused by predicted-LID routing relative to ground-truth LID routing, and demonstrate cross-domain improvements on MASC, Casablanca, and SADA, including gains from fine-tuning Qwen3-ASR on \model.
\end{itemize}

\begin{figure*}[t]
\centering
\includegraphics[width=\linewidth]{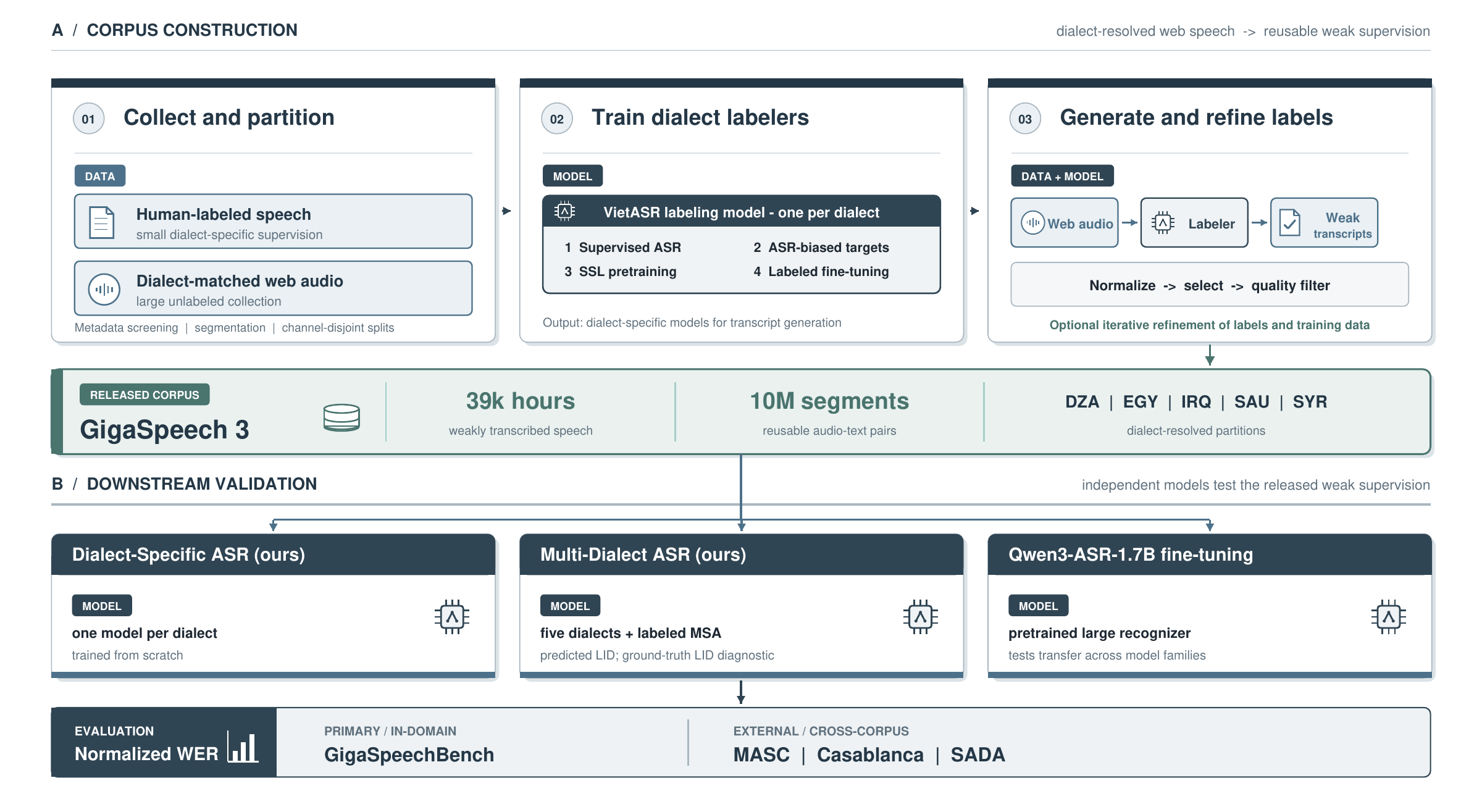}
\caption{Overview of \model{} construction and validation. Channel-disjoint manual and web data train dialect-specific VietASR labeling models, whose hypotheses are filtered to form the five-dialect corpus. Independent Zipformer training and Qwen3-ASR adaptation evaluate the resulting weak supervision; labeled MSA is added only to the Multi-Dialect Zipformer and Qwen training recipes.}
\label{fig:pipeline}
\end{figure*}

\section{Related Work}
\label{sec:related_work}

\subsection{Low-resource ASR through self-supervised and multilingual learning}

Self-supervised speech learning reduces the amount of transcription required to adapt an acoustic model. wav2vec~2.0 learns quantized latent targets with a contrastive objective, while HuBERT predicts clustered hidden units in masked regions \cite{baevski_wav2vec_2020,hsu_hubert_2021}. XLS-R scales this paradigm across languages so that low-resource targets can benefit from cross-lingual pretraining \cite{babu_xls-r_2021}. A complementary line trains a single multilingual recognizer or adapts modular components across languages \cite{pratap_massively_2020,yu_master-asr_2023,ramirez_anatomy_2024}. These approaches improve representations and parameter sharing, but they still require task-specific transcripts during adaptation. Our work uses self-supervised learning to construct dialect-specific labeling models, then converts dialect-matched unlabeled speech into a much larger weakly labeled corpus.

\subsection{ASR-biased speech pretraining}

HuBERT obtains masked-prediction targets by clustering acoustic features and then representations learned in earlier pretraining iterations \cite{hsu_hubert_2021}. A substantial line of work has made these targets more specific to ASR by injecting supervision into their construction \cite{yang2024unifieddiscrete}. Supervision-guided codebooks introduced both PBERT, which uses phoneme alignments from a hybrid recognizer, and CTC clustering, which quantizes representations from a supervised CTC encoder; both recognizers can be trained on the same limited labeled subset used for downstream fine-tuning \cite{wang_supervision_codebooks_2022}. Chen et al. used an ASR encoder to bypass HuBERT's initial acoustic-clustering iteration, reducing the compute required to obtain useful masked-prediction targets \cite{chen_reducing_barriers_2023}. 
Later methods further close the loop between ASR fine-tuning and speech pretraining. ASBERT extracts linguistically informed discrete targets from a fine-tuned ASR encoder and uses them in the next pretraining round \cite{kim_asbert_2023}. Biased SSL likewise fine-tunes the target-generating model toward ASR before reclustering its representations, improving both recognition accuracy and convergence speed \cite{kreyssig_biased_2023}. CTCBERT addresses a related problem at the objective level: it replaces framewise cross-entropy with CTC so that masked prediction is less sensitive to errors in clustered or aligned target boundaries \cite{fan_ctcbert_2023}. Taken together, these approaches show that labeled speech can guide the target generator, the iterative training path, or the pretraining loss.

VietASR builds on this prior line by combining ASR-biased iterative targets with an Fbank front end and a multi-rate Zipformer~\cite{yao2024zipformer} encoder. It starts from an RNN-T model trained on a small labeled set, clusters that encoder's representations, predicts the resulting units on masked Fbank sequences, and fine-tunes the pretrained encoder for ASR. Later iterations repeat target extraction and pretraining with the improved encoder. This design avoids a purely acoustic first-round codebook and adapts the procedure to a compact recognizer that supports online and offline decoding. In the original Vietnamese study, a 68M-parameter model trained with 50 hours of labeled speech was competitive with much larger open and commercial systems \cite{yang2025k2ssl,zhuo2025vietasr}.

We adopt this recipe to obtain five dialect-specific SSL-adapted labeling models; the recipe itself is a foundation rather than the claimed novelty of this work. Our main question begins after those models are trained: whether their transcripts constitute reusable weak supervision for independent downstream models. \model\ evaluates the released corpus through from-scratch Zipformer training and cross-architecture Qwen adaptation. Improvements from independently initialized models provide evidence for the utility of the weakly labeled data beyond the recognizers that generated it.

\subsection{Automatic weak labeling at scale}

Automatic weak labeling uses a recognizer's hypotheses as supervision for otherwise unlabeled audio. Iterative variants refresh those hypotheses as the acoustic model improves, and their effectiveness depends on decoding, augmentation, and control of label errors \cite{xu_iterative_2020}. The approach also scales to multilingual settings: a supervised multilingual model can be adapted to a target language, used to label target-language audio, and followed by a final model trained from scratch or fine-tuned on the generated labels \cite{lugosch_pseudo-labeling_2022}. Large-scale studies on low-resource speech further combine multiple transcription and evaluation models to select useful audio--text pairs \cite{bhogale_empowering_2024}.

GigaSpeech~2 integrates these ideas into a reproducible corpus construction pipeline, including web-audio collection, automatic transcription, alignment, multidimensional filtering, and modified noisy-student training \cite{yang_gigaspeech_2024}. Our corpus construction follows the same broad principle, but the experimental focus differs in two respects. First, compact dialect-specific labeling models trained with ASR-biased SSL generate the Arabic dialect transcripts. Second, we separate the gain obtained from the initial weakly labeled corpus from the additional gain obtained through filtering and iterative refinement. This separation tests whether the initial weak labels are already useful before the more selective refinement stage.

Table~\ref{tab:gigaspeech_evolution} summarizes the technical progression represented by the GigaSpeech name. The original GigaSpeech scaled high-quality supervised English ASR by collecting and validating public audio with available transcripts \cite{chen_gigaspeech_2021}. GigaSpeech~2 shifted the emphasis from existing text pairs to unlabeled web audio for three low-resource Southeast Asian languages, using general-purpose models for transcription, forced alignment, filtering, and iterative refinement \cite{yang_gigaspeech_2024}. \model{} retains large-scale weak labeling but moves the unit of coverage from language to dialect: limited dialect-specific transcripts first adapt the labeling models, and the generated corpus is then validated with independently initialized and pretrained downstream recognizers. Thus, the three corpora share a concern with scalable public-web speech, while differing in language scope, supervision source, and the role of trained models in corpus construction.

\begin{table*}[t]
\centering
\caption{Language coverage, duration, and supervision type in the GigaSpeech series.}
\label{tab:gigaspeech_evolution}
\setlength{\tabcolsep}{6pt}
\renewcommand{\arraystretch}{1.08}
\normalsize
\begin{tabular}{p{0.16\textwidth}p{0.38\textwidth}p{0.14\textwidth}p{0.22\textwidth}}
\toprule[1pt]
Corpus & Language coverage & Duration (h) & Supervision type \\
\midrule
GigaSpeech \cite{chen_gigaspeech_2021}
& English
& 10k
& Human-provided transcripts \\
\addlinespace
GigaSpeech~2 \cite{yang_gigaspeech_2024}
& Thai, Indonesian, and Vietnamese
& 30k
& Refined weak labels \\
\addlinespace
GigaSpeech~3 (ours)
& Arabic: DZA, IRQ, SAU, SYR, and EGY
& 39k
& Refined weak labels \\
\bottomrule[1pt]
\end{tabular}
\end{table*}

YODAS v3 provides a complementary form of web-scale weak supervision: 1.1 million hours of 48-kHz multichannel audio across 147 languages, with platform-provided transcripts when available \cite{chen_yodas3_2026}. Its paper reports roughly 25,000 hours of Arabic using the uploader locale as a language proxy. The current dataset release retains this field as \texttt{locale\_lang}, but the dataset card notes that it is missing for many recordings and organizes language directories by caption-track language when available. Consequently, its partition is broad Arabic rather than a reliable country-level dialect annotation, and the paper does not report Arabic ASR results. \model\ instead curates five dialects, regenerates their labels with dialect-specific recognizers, and evaluates those labels through Arabic dialect ASR models trained from scratch.

\subsection{Arabic and multi-dialect speech recognition}

Spoken Arabic is commonly described through five coarse regional groups: Maghrebi, Egyptian, Levantine, Gulf, and Iraqi or Mesopotamian Arabic \cite{zaidan_arabic_2014}. Maghrebi varieties cover much of North Africa and reflect substantial internal variation and contact with Berber and French; Levantine covers Syria, Lebanon, Jordan, and Palestine; Gulf covers the Arabian Peninsula and Gulf states; Egyptian forms a distinct group; and Iraqi is often treated separately because of its distinctive phonology and morphology, although some taxonomies place it within a broader Gulf group. This hierarchy is only approximate, as variation also occurs by country, city, and social group, but it motivates our selection of DZA, EGY, SYR, SAU, and IRQ as representatives of the five broad groups.

Arabic speech resources have historically concentrated on MSA and a few prominent dialects. MGB-2 released 1,200 hours of lightly supervised Al Jazeera broadcast speech, most of it MSA, with the remaining speech grouped into Egyptian, Gulf, Levantine, and North African varieties \cite{ali_mgb2_2016}. MGB-3 shifted evaluation toward 16 hours of multi-genre Egyptian YouTube speech and paired ASR with five-way identification of Egyptian, Gulf, Levantine, North African, and MSA speech \cite{ali_mgb3_2017}. MGB-5 subsequently introduced Moroccan ASR together with country-level dialect identification data collected from 17 Arabic-speaking countries \cite{ali_mgb-5_2019}. These challenges established increasingly dialect-focused evaluations, but did not provide large-scale ASR training data for country-level varieties.

Larger broadcast resources improve training scale while retaining their source-domain concentration. QASR contains 2,000 hours of lightly transcribed, multi-dialect Al Jazeera speech and associated linguistic metadata\cite{mubarak_qasr_2021}. SADA contributes 668 hours from 57 Saudi TV programs, covering read and spontaneous speech dominated by Saudi varieties with smaller amounts of Yemeni, Egyptian, and Levantine Arabic\cite{alharbi_sada_2024}. Casablanca added human-transcribed conversational speech for eight countries, including Algerian, Egyptian, Emirati, Jordanian, Mauritanian, Moroccan, Palestinian, and Yemeni Arabic\cite{talafha_casablanca_2024}. Together, these corpora broaden domain and regional coverage, but their dialect inventories, scales, and transcription conventions remain unevenly distributed.

Recent manually annotated corpora address complementary gaps in dialect coverage and label granularity. CARDAMOM contains approximately 40 hours of YouTube speech from Egypt, Jordan, Lebanon, Mauritania, Palestine, and Saudi Arabia, with native-speaker transcriptions and labels for 21 sub-country micro-dialects, code-switching, and perceived gender \cite{talafha_cardamom_2026}. It is designed for release as standoff annotations---video IDs, timestamps, transcripts, and metadata with reconstruction code---rather than redistributed audio. Hamsa provides approximately 11.6 hours of similarly focused Emirati speech and shows that dialect-matched fine-tuning improves pretrained Whisper models \cite{alyafeai_hamsa_2026}. These resources prioritize precise gold supervision and fine-grained evaluation; \model{} instead scales country-level weak supervision to thousands of hours per target dialect and tests it with an independently initialized recognizer. CARDAMOM overlaps with our EGY and SAU coverage and resolves within-country varieties, whereas Hamsa supplies a complementary dialect-specific gold benchmark. Their results are therefore complementary rather than directly comparable baselines.

Modeling work has followed three main directions. Dialect-specific adaptation of Whisper and related multilingual models improves covered varieties but generalizes poorly to unseen dialects \cite{talafha_n-shot_2023}. Joint systems share acoustic and lexical knowledge across Arabic varieties, sometimes adding dialect tokens or combining Arabic with English and French to handle code-switching \cite{chowdhury_towards_2021}. Country-level routing has also been studied separately: ADI17 provides roughly 3,000 hours of weakly labeled YouTube speech from 17 countries plus 57 hours of manually refined development and test speech, illustrating both the utility and noise of assigning dialects from source channels \cite{shon_adi17_2020}. Arabic-specific pretraining and weak supervision scale coverage further: ArTST-based studies analyze how pretraining dialect coverage affects generalization, while recent large weakly supervised systems combine broad pretraining with continual supervised fine-tuning \cite{djanibekov_dialectal_2025,salhab_advancing_2025,salhab_munsit_2025}. Across these approaches, gains remain uneven by country and domain, showing that broad Arabic coverage does not remove the need for dialect-matched supervision \cite{wang_open_2024}.

GigaSpeechBench provides the common evaluation used in this work, covering real-world speech across low-resource languages, dialects, accents, domains, and age groups \cite{tu_gigaspeechbench_2026}. We use its released Arabic test sets and keep test channels separate from the channels used to collect training audio. The multi-dialect recognizer uses predicted dialect IDs in the main comparison, and a matched diagnostic with and without ground-truth LID guidance measures the effect of routing errors. Improving dialect identification itself is outside the contribution of this paper.

Arabic orthographic variation also makes evaluation sensitive to text normalization. Surface forms that are linguistically close can produce different WERs under different mappings of letter variants, punctuation, and dialect spellings. We therefore treat the normalizer, test manifest, and decoding condition as part of the experimental protocol. Results produced under incompatible protocols are reported separately and are not used to support direct superiority claims.
\section{Data and Problem Setup}
\label{sec:data}

\subsection{Dialect scope}

The primary study covers five country-level Arabic varieties: Algerian (DZA), Egyptian (EGY), Iraqi (IRQ), Saudi (SAU), and Syrian (SYR). Together they provide representatives of the Maghrebi, Egyptian, Mesopotamian, Gulf, and Levantine groups introduced in Section~\ref{sec:related_work}. This mapping is deliberately coarse: each country contains substantial regional and social variation, and a country tag should not be interpreted as a complete linguistic taxonomy. Our goal is to construct useful dialect-matched supervision at a scale that is currently unavailable for these five evaluation conditions.

MSA is additional model coverage. The Multi-Dialect Zipformer uses the MGB-2 dataset; no MSA-specific VietASR model is trained. The Qwen adaptation study also uses manually labeled MSA speech from MGB-2 \cite{ali_mgb2_2016}. This additional supervision is excluded from the five-dialect weak-label inventory reported below. 

\subsection{Web collection and split integrity}

We collect candidate audio from public web channels and screen sources before model training. The source-level filters use channel identifiers and URLs, title metadata, publication dates, and recording duration to remove irrelevant or unsuitable material. Accepted recordings are segmented into utterance-like regions and pass basic audio-quality checks. A small subset is selected for professional transcription and forms the manually labeled set; the remaining material forms the dialect-specific unlabeled pool.

Separation is enforced at the channel level rather than only at the segment level. Channels assigned to the manually labeled set, the unlabeled collection, and the GigaSpeechBench evaluation data are mutually disjoint. This stronger unit of separation reduces the chance that repeated speakers, recurring programs, or channel-specific acoustic conditions appear on both sides of evaluation. It also makes the corpus-construction claim more demanding: each dialect-specific labeling model must generalize to new channels when generating weak labels.

The collected speech is naturally heterogeneous. It includes differences in recording equipment, background noise, speaking style, topic, and the amount of MSA or code-switching embedded in nominally dialectal content. We preserve this diversity in the large unlabeled pool and rely on dialect-matched labeling models plus downstream selection to construct the final corpus. The resulting task is therefore real-world dialectal speech recognition rather than recognition of a single controlled domain.

\subsection{Supervision inventory}

Table~\ref{tab:pseudo_data} summarizes the supervision available to the five target dialects. The source collections contain 12,183--20,135 hours of unlabeled speech per dialect, whereas only 51.1--63.2 hours per dialect are manually transcribed. The final weak-label inventories contain $\sim$10 million segments and  39,028 hours, ranging from $\sim$4,361 hours for SYR to 11,461 hours for DZA. The released SAU and SYR partitions comprise the filtered weak-label data after four NST rounds.

\begin{table}[t]
\centering
\caption{Training-data duration for the five target dialects (hours).}
\label{tab:pseudo_data}
\fontsize{10.5pt}{12pt}\selectfont
\setlength{\tabcolsep}{2.8pt}
\renewcommand{\arraystretch}{1.15}
\begin{tabular}{lrrr}
\toprule[1pt]
\multirow{2}{*}{Dialect} & \multicolumn{3}{c}{Duration (h)} \\
\cmidrule(lr){2-4}
& Human-labeled & Unlabeled & Refined weak-labeled \\
\midrule
DZA & 61.1 & 17,812 & 11,461 \\
IRQ & 63.2 & 15,531 & 10,162 \\
SAU & 58.3 & 20,135 & 4,512 \\
SYR & 59.0 & 14,905 & 4,361 \\
EGY & 51.1 & 12,183 & 8,533 \\
\midrule
Total & 292.7 & 80,566 & 39,028 \\
\bottomrule[1pt]
\end{tabular}
\end{table}

\subsection{Position relative to existing Arabic resources}

Table~\ref{tab:resource_positioning} separates two directions in Arabic speech-resource development. Casablanca provides human-labeled multidialect training and evaluation data with country labels, CARDAMOM emphasizes fine-grained regional and sociolinguistic annotation, and Hamsa supports Emirati ASR adaptation and evaluation \cite{talafha_casablanca_2024,talafha_cardamom_2026,alyafeai_hamsa_2026}. These resources prioritize transcription quality and linguistic detail at a tens-of-hours scale. YODAS v3 instead provides web-scale multilingual audio and reports roughly 25,000 hours of Arabic, but its \texttt{locale\_lang} metadata are incomplete, and its current \texttt{ar} organization does not supply reliable country-level dialect labels \cite{chen_yodas3_2026}.

\begin{table*}[t]
\centering
\caption{Comparison with recent Arabic speech resources.}
\label{tab:resource_positioning}
\setlength{\tabcolsep}{6pt}
\normalsize
\begin{tabular}{p{0.20\textwidth}p{0.40\textwidth}p{0.12\textwidth}p{0.18\textwidth}}
\toprule[1pt]
Resource & Country codes & Scale (h) & Supervision \\
\midrule
Casablanca \cite{talafha_casablanca_2024} & DZA, EGY, ARE, JOR, MRT, MAR, PSE, YEM & 48 & Human labels \\
CARDAMOM \cite{talafha_cardamom_2026} & EGY, JOR, LBN, MRT, PSE, SAU & 40 & Human labels \\
Hamsa \cite{alyafeai_hamsa_2026} & ARE & 12 & Human labels \\
\midrule
YODAS v3 \cite{chen_yodas3_2026} & Unspecified & 25k & Weak labels \\
\model{} (ours) & DZA, IRQ, SAU, SYR, EGY & 39k & Refined weak labels \\
\bottomrule[1pt]
\end{tabular}
\end{table*}

\model{} occupies a different point in this design space. It uses a few dozen manually transcribed hours to train dialect-specific labeling models, converts much larger web collections into country-level weak supervision, and validates the resulting corpus with independently trained recognizers. The corpus is therefore not a substitute for fine-grained gold corpora; it supplies the scale required for representation learning and robust ASR training while retaining a usable dialect identity.

\section{Method}
\label{sec:method}

\subsection{Overview}

Figure~\ref{fig:pipeline} separates corpus construction into channel-disjoint data preparation, dialect-specific weak-label generation and refinement, and downstream corpus validation. We first collect and partition dialect-matched web speech. A VietASR ASR-biased self-supervised learning procedure converts a small labeled set and a large unlabeled pool into a dialect labeling model. Its transcripts are selected to form an initial weak-label inventory. SAU and SYR then undergo four rounds of noisy-student training (NST), and the final round produces their released inventories and corresponding $r_4$ recognizers. New recognizers test whether the resulting audio--text pairs provide useful supervision beyond the labeling-model parameters.

\subsection{VietASR dialect-labeling model construction}

The labeling-model pipeline follows the four-stage VietASR design \cite{zhuo2025vietasr}. In Stage~1, a labeled-only Zipformer transducer, denoted Iter0, is trained on the dialect's manual transcripts. Although Iter0 is not yet competitive for every dialect, its encoder has learned speech--text alignment cues that are absent from purely acoustic features. Stage~2 runs this supervised encoder on speech and clusters its hidden representations into a discrete codebook. These clusters become ASR-biased targets: they preserve acoustic regularities while being shaped by a recognizer trained on the target dialect.

Stage~3 pretrains a fresh encoder on the large unlabeled pool. Contiguous spans of filter-bank features are masked, and the encoder predicts the cluster identity assigned to each masked frame. For an utterance $X$, masked positions $\mathcal{M}$, and codebook labels $c_t$, the objective is the masked cross-entropy
\begin{equation}
    \mathcal{L}_{\mathrm{SSL}}
    =-\sum_{t\in\mathcal{M}}\log p(c_t\mid\widetilde{X}),
\end{equation}
where $\widetilde{X}$ denotes the masked feature sequence. This retains the computationally efficient Fbank front end and multi-rate Zipformer backbone while steering representation learning toward ASR-relevant distinctions.

In Stage~4, a transducer predictor and joiner are attached to the pretrained encoder, and the full recognizer is fine-tuned on the same manually labeled set. This produces the SSL-adapted labeling model, denoted Iter1, that generates the corpus transcripts. The Iter0--Iter1 comparison holds the quantity of manual supervision fixed and therefore measures the gain obtained by ASR-biased pretraining. VietASR permits later iterations that refresh the codebook from the improved encoder; the labeling models reported here use one SSL-plus-fine-tuning iteration.

\subsection{Weak-label generation and selection}

Each Iter1 labeling model decodes only the dialect-matched web channels that were excluded from the labeled and evaluation sets. The hypotheses are normalized into the same writing space used for downstream training, paired with their audio segments, and screened using duration, recognition score, duplication, and text-quality criteria. The output is a set $\widetilde{\mathcal{D}}_d=\{(x_i,\hat{y}_i)\}$ for dialect $d$, where $\hat{y}_i$ is a model-generated transcript rather than a manual reference.

The distinction between an accurate labeling model and a useful weak-label corpus is central to the experimental design. A labeling model can achieve low WER because of its pretrained parameters even if its generated transcripts are not reusable. We therefore initialize the primary downstream Zipformer independently and optimize it only on $\widetilde{\mathcal{D}}_d$. Improvement over Iter1 under this constraint is evidence that the audio--text pairs contain useful training information beyond the labeling-model weights.

The final corpus contains 10 million segments and 39k hours across the five dialects. DZA, IRQ, and EGY contribute their selected weak-label inventories; SAU and SYR contribute the r4-filtered inventories produced by the refinement procedure below. Selection is dialect-specific: a common target number of hours would hide differences in source composition and label reliability.

\subsection{From-scratch training and iterative refinement}
\label{sec:nst_refinement}

The primary downstream system is a Zipformer2 transducer trained from random initialization on weak labels only. Its predictor and joiner are learned together with the encoder, so neither acoustic representations nor decoding parameters are inherited from Iter1. Dialect-Specific models use one corpus partition and one dialect-specific output vocabulary. This setting is the most direct test of the released corpus because the downstream model can be reproduced from the audio--text pairs without access to the labeling-model checkpoint.

SAU and SYR additionally use GigaSpeech~2-style noisy-student training (NST), which iteratively refines weak supervision and trains successive recognizers \cite{yang_gigaspeech_2024}. We denote the initial weak-label-trained recognizer in each refinement chain by $r_0$. It is the starting point before NST refinement, whereas Iter0 and Iter1 denote the labeled-only and SSL-adapted labeling models described above. These two numbering schemes therefore refer to different stages of corpus construction.

An NST round comprises filtering the current audio--text training pairs, training the next recognizer on the selected weak supervision, and regenerating hypotheses to continue refinement. We use $r_k$ to denote the recognizer after $k$ completed refinement rounds: $r_1$, $r_2$, and $r_3$ are intermediate models, and $r_4$ is the final model after four rounds. A round is a complete refinement-and-training cycle, rather than a single optimization epoch. The round index identifies the model's position in this chain, not a separate model architecture.

The released SAU and SYR partitions are the final, fourth-round-filtered weak-label inventories, containing 4,512 and 4,361 hours, respectively. We refer to these inventories as the r4-filtered data and to the corresponding final recognizers as the NST r4 models. Their WERs are used in the main dialect-specific comparison; Table~\ref{tab:nst_chain} traces their improvement from $r_0$ through $r_4$. We restrict refinement to these two dialects because the adopted recipe assumes a sufficiently accurate labeling model.

\subsection{Multi-dialect training}

The Multi-Dialect Zipformer pools the five weak-label corpora and the MGB-2 dataset. During training, the ground-truth dialect token is prepended to each transcript. The model learns to predict this token followed by the transcript, jointly learning dialect identification and ASR. Dialect tokens extend the shared output vocabulary; training uses no separate LID classification head or LID loss.

At inference, the main evaluation uses predicted IDs to condition recognition. Ground-truth-ID decoding is reserved for a matched diagnostic that measures routing errors. This factorization lets one acoustic model share evidence across varieties while preserving dialect-conditioned output behavior; improving dialect identification itself is outside our scope.

The multi-dialect experiment asks whether the weak-label corpora support one recognizer with competitive aggregate performance and MSA coverage. Differences between the Dialect-Specific and Multi-Dialect columns are not treated as a pure measure of cross-dialect transfer because the models also differ in data mixture, vocabulary, conditioning, and automatic-routing errors.

\subsection{Cross-architecture transfer}

Finally, we fine-tune Qwen3-ASR-1.7B using weak transcriptions from our five dialect partitions together with manually labeled MSA speech from MGB-2, and evaluate the adapted model without and with ground-truth LID guidance at decoding. These are decoding conditions, not separate claims about training with or without dialect conditioning. This model is substantially larger and already pretrained, so it is not the primary controlled validation. Instead, it tests portability: if the generated transcripts improve both an independently initialized Zipformer and a pretrained large recognizer, their value is less likely to depend on one architecture or optimization recipe.

\section{Experimental Setup}
\label{sec:setup}

\subsection{Evaluation questions}

The experiments are organized around five questions. \textbf{RQ1} asks whether ASR-biased SSL improves a dialect labeling model when manual supervision is fixed. \textbf{RQ2} asks whether a new Zipformer trained only on weak labels can improve further, thereby validating the generated corpus independently of the labeling-model parameters. \textbf{RQ3} asks whether five dialect corpora can train one dialect-conditioned recognizer that remains competitive on every variety and also covers MSA. \textbf{RQ4} asks whether the weak labels remain useful under iterative refinement and when transferred to Qwen3-ASR-1.7B. \textbf{RQ5} asks whether the resulting models generalize to independently released Arabic corpora rather than benefiting only from the GigaSpeechBench evaluation domain.

These questions determine the comparison chain. Iter0 uses only manual transcripts. Iter1 adds SSL on the corresponding unlabeled collection and then fine-tunes on the same transcripts. The weak-only system discards the Iter1 parameters and trains on generated labels from scratch. The multi-dialect and Qwen experiments broaden the model class after the dialect-specific resource has been validated.

\subsection{Recognition systems}

All primary recognizers use the icefall Zipformer2 transducer recipe. Dialect-Specific Zipformers use dialect-specific 500-unit BPE tokenizers where recorded, whereas the Multi-Dialect Zipformer uses a shared 1,500-unit BPE vocabulary and automatically predicted LID at decoding. Final Dialect-Specific SAU/SYR results use the NST r4 models corresponding to their r4-filtered released inventories. The multi-rate Zipformer encoder is paired with a stateless predictor and transducer joiner. Checkpoint averaging and decoding follow the settings recorded in the experiment registry.

The Qwen3-ASR-1.7B study uses the released base recognizer and two decoding conditions for the adapted model: without and with ground-truth LID guidance. Fine-tuning combines weakly transcribed dialect speech with manually labeled MSA data from MGB-2; the MSA condition therefore does not test weak supervision alone. QwenCleo\footnote{\url{https://github.com/MohammedAly22/qwencleo-asr}} and Whisper-CS\footnote{\url{https://huggingface.co/MohamedRashad/Arabic-Whisper-CodeSwitching-Edition}} provide contextual Arabic-specialized large-model baselines, but they are not data-matched controls.

\subsection{Test sets and normalization}

We use the released GigaSpeechBench Arabic test sets as the primary evaluation \cite{tu_gigaspeechbench_2026}. The test audio and reference transcripts are identical for the proposed and off-the-shelf comparisons, and all hypotheses pass through the benchmark's common Arabic normalizer. Historical exploratory runs that used a different manifest or normalization path are excluded from the reported comparison.

To test whether the conclusions depend on this single benchmark domain, we additionally evaluate on subsets of three independently released Arabic corpora: SADA \cite{alharbi_sada_2024}, Casablanca \cite{talafha_casablanca_2024}, and MASC \cite{alfetyani_masc_2023}. We retain the target varieties represented in our training data: DZA and EGY from Casablanca; DZA, IRQ, SAU, SYR, and EGY from MASC; and the SAU-targeted SADA evaluation set. For each corpus, its official development and test partitions are concatenated into one evaluation pool. Neither partition is used for model selection, hyperparameter tuning, checkpoint selection, or text-normalization design. Table~\ref{tab:external_eval_data} summarizes the duration of the primary GigaSpeechBench test sets and the three external evaluation corpora by dialect.

\begin{table}[t]
\centering
\caption{Duration of the Arabic evaluation sets by dialect (hours). Development and test partitions are merged for MASC, Casablanca, and SADA.}
\label{tab:external_eval_data}
\setlength{\tabcolsep}{3.5pt}
\renewcommand{\arraystretch}{1.15}
\normalsize
\begin{tabular}{@{}lrrrrr@{}}
\toprule[1pt]
\multirow{2}{*}{Dataset} & \multicolumn{5}{c}{Duration (h)} \\
\cmidrule(lr){2-6}
& DZA & IRQ & SAU & SYR & EGY \\
\midrule
GigaSpeechBench & 20.0 & 20.4 & 20.3 & 20.2 & 20.5 \\
MASC & 0.1 & 0.7 & 1.4 & 6.6 & 4.5 \\
Casablanca & 1.9 & -- & -- & -- & 2.0 \\
SADA & -- & -- & 19.9 & -- & -- \\
\midrule
Total & 22.0 & 21.0 & 41.6 & 26.8 & 27.0 \\
\bottomrule[1pt]
\end{tabular}
\end{table}

For the external evaluation suite, we use the same final checkpoints as in the primary comparison, including the SAU/SYR NST r4 models. The Multi-Dialect Zipformer and fine-tuned Qwen use automatically predicted LID without ground-truth LID guidance. Decoding settings are fixed before evaluation, and the same Arabic text normalizer is applied to references and hypotheses. Results are reported for each corpus--variety pair instead of only after pooling all utterances. This prevents the much larger SADA and MASC-SYR subsets from obscuring performance on smaller conditions; in particular, the three-minute MASC-DZA subset is retained for completeness but is not treated as strong standalone evidence.

Word error rate (WER) is the primary metric and is reported as a percentage. The main system comparison averages DZA, EGY, IRQ, SAU, and SYR; the LID diagnostic and Qwen adaptation tables average these five dialects plus MSA. The external evaluation averages its eight corpus--variety conditions. All averages are unweighted means of the listed WERs, not pooled word-error rates, and should be compared only within the same coverage.

\subsection{External systems and dialect-ID diagnostic}

The third-party values in Table~\ref{tab:third_party} are reproduced from the common GigaSpeechBench v3 evaluation table. The comparison includes commercial APIs, general multilingual recognizers, and publicly released Arabic-capable models. We use the strongest complete system as the principal external reference and italicize the best third-party result for each dialect; bold and underlining mark the best and second-best values overall. This avoids constructing a single baseline by selecting different systems after observing each test set.

The reported Multi-Dialect Zipformer is obtained by averaging checkpoints from epochs 4--8 and uses automatically predicted variety tags in the main GigaSpeechBench comparison. Using this same averaged model, a paired diagnostic compares decoding without and with ground-truth LID guidance and reports the classifier's confusion matrix. Its WERs characterize routing cost under a matched model and test set. The classifier is an inference component; dialect-identification methodology is not claimed as a contribution.

The duration-stratified LID analysis uses the same averaged model with greedy decoding and predicted IDs. During training, the ground-truth dialect token is prepended to the transcript so that LID and ASR are learned jointly under the transducer objective. We group utterances into durations below 3 seconds, from 3 to less than 10 seconds, and at least 10 seconds. Within each variety and bucket, LID accuracy is the fraction assigned to the reference variety, and sample counts accompany the accuracies to show the evaluation support.

\section{Results and Analysis}
\label{sec:results}

\subsection{From manual labels to reusable weak supervision}

Table~\ref{tab:main_results} addresses RQ1 and RQ2 by comparing the labeling models with the final corpus-trained systems. ASR-biased SSL followed by labeled fine-tuning improves every dialect over its labeled-only Iter0 model. The relative WER reductions are 28.1\% for DZA, 26.2\% for IRQ, 27.6\% for EGY, and 74.3\% for SYR. SAU starts from a much stronger 13.17 WER and improves by a smaller 1.4\% relative. The spread shows that SSL is most valuable where the few-dozen-hour supervised baseline remains weak.

\begin{table*}[t]
\centering
\caption{GigaSpeechBench WER (\%) by model and dialect. The final models use the released weak transcriptions; SAU and SYR include four-round NST refinement.}
\label{tab:main_results}
\normalsize
\setlength{\tabcolsep}{5.2pt}
\renewcommand{\arraystretch}{1.2}
\begin{tabular}{lrrrrrr}
\toprule[1pt]
Model & DZA & IRQ & SAU & SYR & EGY & MSA \\
\midrule
Labeled only & 53.47 & 42.50 & 13.17 & 58.88 & 43.92 & 15.98 \\
SSL + FT & 38.47 & 31.35 & 12.98 & 15.11 & 31.81 & -- \\
Dialect-Specific & \textbf{33.64} & 29.99 & \textbf{12.12} & \textbf{14.25} & \textbf{29.32} & -- \\
Multi-Dialect & 36.78 & \textbf{28.85} & 12.81 & 16.33 & 32.80 & \textbf{10.83} \\
\bottomrule[1pt]
\end{tabular}
\end{table*}

The final Dialect-Specific Zipformers improve over their corresponding Iter1 labeling models on all five dialects. DZA, IRQ, and EGY reach 33.64, 29.99, and 29.32 WER, respectively, corresponding to relative reductions of 12.6\%, 4.3\%, and 7.8\%. SAU and SYR reach 12.12 and 14.25 after four NST rounds, improving over Iter1 by 6.6\% and 5.7\%. Their released 4,511.62-hour and 4,360.75-hour inventories are the r4-filtered weak-label data. These two final results therefore measure the full construction-and-refinement pipeline; the round-wise analysis below isolates their NST gains.

VietASR and downstream corpus training address complementary bottlenecks. VietASR uses limited manual transcripts to build an accurate labeling model; weak labeling then creates much larger audio--text training sets for new recognizers. The downstream models learn from these pairs without consuming the manual transcripts or inheriting Iter1's SSL-adapted parameters. Their performance supports the corpus as reusable supervision, while the SAU/SYR refinement chains account for the additional processing used in those two releases.

\subsection{Comparison with third-party systems}

Table~\ref{tab:third_party} compares systems with complete results on the five target dialects in GigaSpeechBench v3. The strongest complete third-party system is Qwen3.5-Omni-Plus, with a 28.62 macro-average WER. Our final Dialect-Specific Zipformers reach 23.86, a 16.6\% relative reduction. The automatically routed Multi-Dialect Zipformer reaches 25.51, a 10.9\% reduction, while ground-truth LID guidance lowers its diagnostic result to 23.62. The dialect-specific aggregate includes the NST r4 models for SAU and SYR.

\begin{table*}[t]
\centering
\caption{WER (\%) on five GigaSpeechBench v3 dialects \cite{tu_gigaspeechbench_2026}. Third-party systems are ordered by decreasing average WER; italics mark the best third-party value, and bold/underlining mark the best/second-best overall.}
\label{tab:third_party}
\normalsize
\setlength{\tabcolsep}{5pt}
\renewcommand{\arraystretch}{1.2}
\begin{tabular}{lrrrrrr}
\toprule[1pt]
Model & DZA & IRQ & SAU & SYR & EGY & Avg. \\
\midrule
Dolphin Base & 78.26 & 65.20 & 52.35 & 38.12 & 85.31 & 63.85 \\
Dolphin Small & 72.44 & 62.05 & 50.91 & 30.03 & 74.70 & 58.03 \\
GPT-4o Transcribe & 63.14 & 54.53 & 42.38 & 31.67 & 64.23 & 51.19 \\
FunASR-Realtime & 66.30 & 53.44 & 37.67 & 24.24 & 63.33 & 49.00 \\
Whisper Large~v3 & 72.02 & 51.04 & 32.79 & 19.12 & 69.78 & 48.95 \\
Deepgram Nova~3 & 57.90 & 47.54 & 25.02 & 30.61 & 52.77 & 42.77 \\
NVIDIA NeMo & 62.66 & 43.22 & 29.28 & 20.13 & 54.83 & 42.02 \\
Qwen3-ASR-1.7B & 63.43 & 41.27 & 25.85 & 18.50 & 59.23 & 41.66 \\
Meta OmniASR~3B & 57.68 & 38.80 & 25.31 & 17.86 & 52.37 & 38.40 \\
ElevenLabs Scribe~v2 & 50.43 & 38.67 & 33.33 & 14.73 & 44.44 & 36.32 \\
Qwen3-ASR-Flash & 57.18 & 33.21 & 19.21 & 14.41 & 48.78 & 34.56 \\
Chirp~3 & 53.11 & 35.71 & 16.76 & 24.13 & 42.71 & 34.48 \\
Azure & 51.22 & 34.61 & 20.09 & 17.74 & 47.65 & 34.26 \\
Gemini 3.0 Flash & \textit{44.22} & 36.55 & 20.10 & 14.40 & 41.22 & 31.30 \\
Qwen3.5-Omni-Plus & 47.11 & \underline{\textit{28.54}} & \textit{16.56} & \underline{\textit{13.76}} & \textit{37.12} & \textit{28.62} \\
\midrule
Dialect-Specific (ours) & \textbf{33.64} & 29.99 & \textbf{12.12} & 14.25 & \underline{29.32} & \underline{23.86} \\
Multi-Dialect w/o LID (ours) & 36.78 & 28.85 & 12.81 & 16.33 & 32.80 & 25.51 \\
Multi-Dialect w/ LID (ours) & \underline{35.29} & \textbf{27.84} & \underline{12.62} & \textbf{13.18} & \textbf{29.16} & \textbf{23.62} \\
\bottomrule[1pt]
\end{tabular}
\end{table*}

The final Dialect-Specific Zipformers outperform the best third-party values on DZA, SAU, and EGY by 23.9\%, 26.8\%, and 21.0\%, respectively. Under automatic routing, Qwen3.5-Omni-Plus remains stronger on IRQ and SYR at 28.54 and 13.76, compared with 28.85 and 16.33 for the Multi-Dialect model. Ground-truth LID guidance lowers the results to 27.84 and 13.18 on these two dialects, showing that routing errors account for part of the remaining gap. The practical automatically routed system nevertheless improves substantially over the strongest complete third-party macro-average.

The primary systems are compact Zipformer transducers trained from the constructed corpus, whereas several baselines are large pretrained or commercial recognizers. Their lower macro-average WER shows the value of dialect-matched weak supervision for compact ASR. The remaining IRQ and SYR gaps also show that corpus scale and specialization do not guarantee superiority on every dialect.

\subsection{Cross-corpus validation}

Table~\ref{tab:external_results} answers RQ5 on eight corpus--variety conditions from MASC, Casablanca, and SADA. A \model-trained Zipformer obtains the lowest WER among the six evaluated systems in every condition. The Multi-Dialect Zipformer is strongest on four of the five MASC subsets and reduces the MASC macro-average from 32.64 for the Dialect-Specific Zipformers to 27.21. The Dialect-Specific Zipformers are stronger on MASC-SAU, both Casablanca subsets, and SADA. Across all eight conditions, their macro-average WERs are 34.39 and 31.95, respectively.

\begin{table*}[t]
\centering
\caption{Cross-corpus WER (\%) on MASC, Casablanca, and SADA. The same final checkpoints are used throughout; ``Avg.'' is the unweighted mean over eight corpus--variety conditions. Bold and underlining mark the best and second-best configurations.}
\label{tab:external_results}
\normalsize
\setlength{\tabcolsep}{3pt}
\renewcommand{\arraystretch}{1.15}
\begin{tabular*}{\textwidth}{@{\extracolsep{\fill}}lrrrrrrrrr@{}}
\toprule[1pt]
\multirow{2}{*}{Model} & \multicolumn{5}{c}{MASC} & \multicolumn{2}{c}{Casablanca} & SADA & \\
\cmidrule(lr){2-6}\cmidrule(lr){7-8}\cmidrule(lr){9-9}
& DZA & IRQ & SAU & SYR & EGY & DZA & EGY & SAU & Avg. \\
\midrule
Whisper-CS & 59.49 & 51.96 & 45.05 & 28.43 & 41.45 & 90.52 & 51.85 & 60.39 & 53.64 \\
QwenCleo & 62.26 & 51.64 & 47.07 & 29.48 & 35.74 & 88.12 & 49.08 & 50.67 & 51.76 \\
Qwen3-ASR-1.7B (Base) & 56.08 & 41.79 & 34.81 & 22.51 & 40.80 & 82.89 & 57.83 & 40.06 & 47.10 \\
\midrule
Qwen3-ASR-1.7B FT w/o LID (ours) & 48.61 & 36.03 & 24.03 & 20.95 & 36.29 & 62.59 & 42.76 & 30.97 & 37.78 \\
Dialect-Specific (ours) & 49.57 & 37.01 & \textbf{21.23} & 21.33 & 34.04 & \textbf{50.06} & \textbf{37.34} & \textbf{24.54} & 34.39 \\
Multi-Dialect w/o LID (ours) & \textbf{40.17} & \textbf{30.46} & 23.08 & \textbf{15.44} & \textbf{26.88} & 51.37 & 41.29 & 26.87 & \textbf{31.95} \\
Multi-Dialect w/ LID (ours) & \underline{43.59} & \underline{32.72} & \underline{22.81} & \underline{17.53} & \underline{29.47} & \underline{50.71} & \underline{37.76} & \underline{26.48} & \underline{32.63} \\
\bottomrule[1pt]
\end{tabular*}
\end{table*}

The two Zipformer configurations are complementary rather than uniformly ordered across corpora. Multi-dialect training helps most on MASC-DZA, IRQ, SYR, and EGY, whereas dialect-specific training transfers better to Casablanca and SADA. This reversal indicates that the benefit of pooling depends on corpus and variety, and motivates reporting each condition instead of a single pooled WER. The very small MASC-DZA subset should also be interpreted cautiously.

Ground-truth LID guidance does not provide consistent additional gains on these out-of-domain sets: the eight-condition macro-average is 32.63, compared with 31.95 under predicted-LID decoding. Most degradations occur on MASC, whose dialect subsets are relatively small, particularly DZA. This pattern may be related to limited evaluation support and domain-specific calibration of the LID component, and requires further study on larger out-of-domain sets.

The Qwen comparison provides a model-family-controlled test of corpus utility. Fine-tuning with \model{} lowers WER relative to the Qwen3-ASR-1.7B base model in all eight conditions, with relative reductions ranging from 6.9\% to 31.0\%. Its macro-average falls from 47.10 to 37.78, a 19.8\% relative reduction. The adapted model also outperforms Whisper-CS in all eight conditions and QwenCleo in seven, with substantially lower macro-averages than both contextual baselines (53.64 and 51.76, respectively). Together with the Zipformer results, this cross-corpus consistency makes it less likely that the primary gains arise only from matching the GigaSpeechBench evaluation domain. We do not claim state of the art on these external corpora because the table compares only six models decoded under our fixed protocol.

\subsection{Dialect-specific versus multi-dialect training}

With automatically predicted LID, the Multi-Dialect Zipformer improves over the final Dialect-Specific Zipformer on IRQ by 1.14 WER points. The Dialect-Specific Zipformers are stronger on DZA, EGY, SYR, and SAU by 3.14, 3.48, 2.08, and 0.69 points, respectively. The multi-dialect system therefore trades some dialect-specific accuracy for unified coverage, with a five-dialect macro-average of 25.51 compared with 23.86 for the dialect-specific models. It also recognizes MSA at 10.83 WER using MGB-2 training data.

This comparison is not a controlled measurement of cross-dialect transfer. The Dialect-Specific and Multi-Dialect Zipformers differ in vocabulary, sampling mixture, MSA coverage, and dialect conditioning; the dialect-specific SAU/SYR models also use NST refinement. Automatic routing introduces an additional source of errors for the multi-dialect model. The practical conclusion is that the corpus supports both dialect-specific recognizers and one automatically routed multi-dialect recognizer, while the paired LID diagnostic below measures routing cost at a fixed checkpoint.

\subsection{Effect of ground-truth LID guidance}

The last two rows of Table~\ref{tab:third_party} isolate the cost of automatic routing using the Multi-Dialect Zipformer. Across the five target dialects, predicted LID raises macro-average WER from 23.62 to 25.51. EGY and SYR incur the largest increases, at 3.64 and 3.15 points, whereas SAU changes by only 0.19 points. The ``w/o LID'' row uses predicted LID and is the practical Multi-Dialect condition; the paired ``w/ LID'' row changes only the LID guidance.

The condition with ground-truth LID guidance estimates performance when routing errors are removed. Table~\ref{tab:lid_confusion} characterizes the errors under automatic routing. MSA is identified almost perfectly, but only 58--74\% of the dialectal utterances are classified correctly. A large share of DZA, EGY, and SAU is routed to MSA, while SYR is additionally confused with IRQ. Together with the WER results, these errors identify routing as a practical bottleneck; LID accuracy alone does not determine the size of the recognition penalty.

\begin{table*}[t]
\centering
\caption{LID confusion matrix for the Multi-Dialect Zipformer averaged over epochs 4--8. Each cell gives the number of utterances followed by its row percentage; bold diagonal entries are correct predictions.}
\label{tab:lid_confusion}
\normalsize
\setlength{\tabcolsep}{7.0pt}
\renewcommand{\arraystretch}{1.15}
\begin{tabular}{lrrrrrrr}
\toprule[1pt]
True $\backslash$ Pred. & DZA & IRQ & SAU & SYR & EGY & MSA & Acc. \\
\midrule
DZA & \textbf{10707 (64\%)} & 44 (0\%) & 348 (2\%) & 18 (0\%) & 557 (3\%) & 4985 (30\%) & 64.3\% \\
IRQ & 78 (1\%) & \textbf{10241 (74\%)} & 601 (4\%) & 127 (1\%) & 581 (4\%) & 2249 (16\%) & 73.8\% \\
SAU & 142 (1\%) & 663 (6\%) & \textbf{6757 (58\%)} & 119 (1\%) & 465 (4\%) & 3443 (30\%) & 58.3\% \\
SYR & 216 (3\%) & 909 (11\%) & 556 (7\%) & \textbf{5004 (60\%)} & 91 (1\%) & 1583 (19\%) & 59.9\% \\
EGY & 27 (0\%) & 205 (1\%) & 636 (4\%) & 44 (0\%) & \textbf{10159 (63\%)} & 4961 (31\%) & 63.4\% \\
MSA & 6 (0\%) & 14 (0\%) & 14 (0\%) & 7 (0\%) & 15 (0\%) & \textbf{4529 (99\%)} & 98.8\% \\
\bottomrule[1pt]
\end{tabular}
\end{table*}

\begin{table}[t]
\centering
\caption{Duration-stratified LID accuracy for the Multi-Dialect model averaged over epochs 4--8. Each cell reports accuracy (\%) followed by the number of utterances in parentheses.}
\label{tab:lid_duration}
\small
\setlength{\tabcolsep}{3pt}
\renewcommand{\arraystretch}{1.2}
\begin{tabular}{@{}lccc@{}}
\toprule[1pt]
Variety & $t < 3$ s & $3 \leq t < 10$ s & $t \geq 10$ s \\
\midrule
DZA & 39.8 (7,808) & 85.0 (8,300) & 99.1 (551) \\
IRQ & 51.5 (4,981) & 84.6 (7,620) & 96.6 (1,276) \\
SAU & 8.4 (2,801) & 76.0 (7,436) & 64.4 (1,352) \\
SYR & 24.4 (394) & 61.7 (5,037) & 61.4 (2,928) \\
EGY & 65.6 (7,030) & 62.3 (8,252) & 54.1 (750) \\
MSA & 97.9 (192) & 98.8 (4,385) & 100.0 (8) \\
\bottomrule[1pt]
\end{tabular}
\end{table}

\noindent\textbf{Duration-dependent confusion.} Table~\ref{tab:lid_duration} provides a complementary duration-stratified analysis of the Multi-Dialect Zipformer's LID predictions. Short utterances are a difficult condition for several dialects: accuracy rises from 39.8\% below 3 seconds to 85.0\% at 3--10 seconds for DZA, from 51.5\% to 84.6\% for IRQ, and from 8.4\% to 76.0\% for SAU. The short-utterance confusion matrices show a strong tendency to predict MSA: 69.3\% of short SAU utterances, 62.2\% of short SYR utterances, and 52.4\% of short DZA utterances receive that label. This pattern is consistent with limited dialect-discriminative evidence in short speech, although the analysis does not establish whether pooling, training composition, or linguistic overlap causes these errors.

Longer speech does not uniformly resolve dialect confusion. DZA and IRQ improve further above 10 seconds, whereas SAU peaks in the middle bucket and EGY accuracy decreases from 65.6\% to 54.1\% across the three buckets. SYR remains near 61\% in both the middle and long buckets, so utterance duration alone does not explain its aggregate weakness. These are associations across different utterances, rather than a controlled test of increasing the duration of the same speech. They also do not directly quantify ASR degradation: SAU has poor short-utterance LID but only a 0.19-point aggregate WER difference between the two Multi-Dialect rows in Table~\ref{tab:third_party}. Paired WER measurements with and without ground-truth LID guidance within each duration bucket would be needed to localize the recognition cost.

These results separate two claims that are easy to conflate. Ground-truth LID guidance improves average WER relative to predicted-LID decoding in the paired Zipformer diagnostic, but reliable automatic dialect identification remains unresolved. This experiment does not compare dialect conditioning against a recognizer with no dialect input. Improving LID is model- and taxonomy-dependent and is outside our scope.

\subsection{Iterative refinement}

Table~\ref{tab:nst_chain} follows each noisy-student chain from its initial weak-label-trained recognizer ($r_0$) to the final recognizer after four refinement rounds ($r_4$), using the round definitions in Section~\ref{sec:nst_refinement}. Under a fixed evaluation protocol, WER decreases at every round: SAU improves from 13.13 to 12.12 and SYR from 14.95 to 14.25. These are absolute reductions of 1.01 and 0.70 WER points, or 7.7\% and 4.7\% relative, respectively. The largest single-round gains occur at different stages: $r_1$ to $r_2$ for SAU and $r_0$ to $r_1$ for SYR.

\begin{table*}[t]
\centering
\caption{Noisy-student refinement on SAU and SYR (WER \%). $r_0$ is the initial weak-label model and $r_1$--$r_4$ are successive refinements; released data correspond to $r_4$.}
\label{tab:nst_chain}
\normalsize
\setlength{\tabcolsep}{5pt}
\renewcommand{\arraystretch}{1.05}
\begin{tabular}{llllllr}
\toprule[1pt]
Dialect & $r_0$ (initial) & $r_1$ & $r_2$ & $r_3$ & $r_4$ (final) & Rel. reduction \\
\midrule
SAU & 13.13 & 12.96 & 12.42 & 12.32 & \textbf{12.12} & 7.7\% \\
SYR & 14.95 & 14.69 & 14.51 & 14.29 & \textbf{14.25} & 4.7\% \\
\bottomrule[1pt]
\end{tabular}
\end{table*}

The r4 endpoints are the final SAU and SYR models used in the main comparison, and the released weak-label inventories for these dialects are the corresponding r4-filtered data. The round-wise results distinguish gains from refinement from those of initial weak supervision: the 7.7\% and 4.7\% reductions are measured relative to each chain's $r_0$ model, rather than to Iter1. In particular, SAU's $r_0$ WER of 13.13 is slightly higher than Iter1's 12.98; the all-five-dialect improvement claim applies to the final models, not uniformly to the initial weak-label models. The release therefore combines initially selected data for DZA, IRQ, and EGY with four-round-refined data for SAU and SYR.

\subsection{Cross-architecture transfer}

Table~\ref{tab:qwen_transfer} addresses RQ4 with a pretrained 1.7B-parameter recognizer. Decoding the adapted model without ground-truth LID guidance reduces WER on all six reported varieties by 13.9--38.3\% relative to the base model, lowering macro-average WER from 38.42 to 28.28 (26.4\% relative). With ground-truth LID guidance, the reductions range from 25.3\% to 44.4\%, and the macro-average falls further to 24.10 (37.3\% relative to the base model). Ground-truth LID guidance improves every reported dialect over predicted-LID routing, with the largest absolute reductions on DZA and EGY. This consistent routing gain indicates that LID errors remain an important bottleneck for the adapted multi-dialect recognizer.

\begin{table*}[t]
\centering
\caption{Large-model comparison on GigaSpeechBench (WER \%). Fine-tuned Qwen uses weakly labeled dialect speech and manually labeled MGB-2 speech for MSA; Avg. is the unweighted macro-average.}
\label{tab:qwen_transfer}
\setlength{\tabcolsep}{5pt}
\normalsize
\renewcommand{\arraystretch}{1.05}
\begin{tabular}{lrrrrrrr}
\toprule[1pt]
Model & DZA & IRQ & SAU & SYR & EGY & MSA & Avg. \\
\midrule
QwenCleo & 71.04 & 55.46 & 32.67 & 28.51 & 44.65 & 24.79 & 42.85 \\
Whisper-CS & 74.86 & 57.86 & 31.02 & 28.24 & 58.91 & 26.56 & 46.24 \\
\midrule
Qwen3-ASR-1.7B (Base) & 63.43 & 41.27 & 25.85 & 18.50 & 59.23 & 22.22 & 38.42 \\
Qwen3-ASR-1.7B FT w/o LID & 47.71 & 34.48 & 16.81 & 15.93 & 41.02 & 13.72 & 28.28 \\
Qwen3-ASR-1.7B FT w/ LID & \textbf{39.25} & \textbf{30.47} & \textbf{14.63} & \textbf{13.82} & \textbf{32.91} & \textbf{13.53} & \textbf{24.10} \\
\bottomrule[1pt]
\end{tabular}
\end{table*}

The adapted Qwen variants outperform QwenCleo and Whisper-CS on every reported dialect, although those public fine-tunes are contextual rather than data-matched controls. More significant for our argument is the direction of transfer: adaptation using weakly labeled dialect data improves Qwen on all five target dialects, complementing the final Zipformer results. The MSA improvement comes from an adaptation setting that includes manually labeled MGB-2 speech and is not evidence for weak-label-only MSA training. The dialect results reduce the likelihood that the corpus is useful only because it matches one Zipformer training recipe.

\section{Discussion}
\label{sec:discussion}

\subsection{What the experiments establish}

Independently initialized recognizers trained on the generated audio--text pairs outperform Iter1 on all five dialects. DZA, IRQ, and EGY use selected weak labels, while SAU and SYR use r4-filtered data. The round-wise experiment isolates refinement gains; external evaluation and Qwen adaptation demonstrate utility beyond GigaSpeechBench and one model family. VietASR supplies labeling models, while independent training tests the corpus beyond those models' parameters.

\subsection{Scale, diversity, and label noise}

Weak labels exchange transcription accuracy for coverage. Selection controls noisy hypotheses, while thousands of hours of dialect-matched speech provide diversity unavailable in the 51--63-hour manually transcribed sets. Volume alone does not explain the results: SAU begins with the largest pool but retains the smallest fraction after refinement, and SYR has the largest SSL gain despite similar retention. Labeling accuracy, source match, diversity, and residual noise remain intertwined; controlled scaling and quality-stratification experiments are needed to separate their effects.

The public-system comparison also bounds the performance claim. The final Dialect-Specific Zipformers lead on DZA, EGY, and SAU, whereas Qwen3.5-Omni-Plus remains stronger than automatic routing on IRQ and SYR. Ground-truth LID guidance narrows these gaps, showing that routing explains part of the difference.

Cross-corpus transfer is not uniform: the Multi-Dialect Zipformer is stronger on four MASC subsets, whereas the Dialect-Specific models lead on both Casablanca subsets and SADA. Pooled training and specialization therefore respond differently to domain shift. Given the six-model comparison and three-minute MASC-DZA subset, these results are evidence of generalization rather than a state-of-the-art claim.

\subsection{Dialect conditioning as an interface}

The Multi-Dialect model predicts a dialect token followed by the transcript under the same transducer objective. Its practical results include automatic LID errors; with the same averaged checkpoint, ground-truth IDs reduce five-dialect macro-average WER from 25.51 to 23.62. Source metadata may supply IDs in curated settings, whereas open-audio routing remains unresolved and should be evaluated separately from ASR.

Duration-stratified results reinforce the value of retaining source metadata: short dialectal speech is often routed to MSA, yet longer inputs do not remove all confusion. Country-level tags also collapse within-country varieties, speaker communities, code-switching, and orthographic choices. Aggregate LID accuracy should therefore be interpreted with duration-specific behavior, and gold resources such as CARDAMOM, Casablanca, and Hamsa remain important for finer evaluation and adaptation.

\section{Corpus Release Plan and Limitations}
\label{sec:limitations}

The \model\ corpus, weak transcripts, and metadata are released for research and evaluation under a research-only license. We also provide construction and training recipes and an online demo; code uses a separate software license. The corpus includes selected audio, weak transcripts, dialect labels, and segment metadata, subject to source-platform terms.

We also provide a community proofreading platform where volunteers can listen to released segments and propose transcript corrections for future corpus revisions; the current release has not been exhaustively human-verified.

The study has four main limitations: automatic dialect-routing errors affect multi-dialect ASR; WER conflates recognition errors with legitimate spelling variation; weak-label quantity, confidence, and source diversity are not independently controlled; and web data may inherit demographic, topical, speaker, and transcription biases. Numerical expressions and code-switching also remain limited. Future extensions should add targeted data and evaluation that separates recognition from formatting errors and checks preservation of foreign-language content.

\section{Conclusion}
\label{sec:conclusion}

We presented \model, a 39K-hour weakly labeled corpus for Arabic dialect ASR. ASR-biased SSL, selection, and NST refinement turn web speech into reusable supervision. Final Dialect-Specific Zipformers improve over their labeling models on all five dialects and reach 23.86 macro-average WER on GigaSpeechBench, 16.6\% below the strongest complete third-party system. A Multi-Dialect model with predicted LID reaches 25.51 and supports MSA. Across eight external conditions, a \model-trained Zipformer leads the six evaluated models, while corpus adaptation consistently improves Qwen3-ASR. These results demonstrate utility across domains and model families and identify automatic dialect routing as a remaining challenge. We thank the Hugging Face team for supporting the hosting and public release of the GigaSpeech 3 corpus.

\section{Ethics Statement}

All collected audio is sourced from materials released under a Creative Commons license. Personally identifiable information has been anonymized using rule-based scripts to remove identifiable content from the data. All annotators are compensated fairly by a professional data annotation company. GigaSpeech 3 is released for non-commercial research and education purposes only, under the dataset license, with ongoing maintenance for identified risks.

\bibliographystyle{IEEEtran}
\bibliography{ref}

\end{document}